\def\year{2022}\relax
\documentclass[letterpaper]{article} % DO NOT CHANGE THIS
\usepackage{aaai22}  % DO NOT CHANGE THIS
\usepackage{times}  % DO NOT CHANGE THIS
\usepackage{helvet}  % DO NOT CHANGE THIS
\usepackage{courier}  % DO NOT CHANGE THIS
\usepackage[hyphens]{url}  % DO NOT CHANGE THIS
\usepackage{graphicx} % DO NOT CHANGE THIS
\usepackage{natbib}  % DO NOT CHANGE THIS AND DO NOT ADD ANY OPTIONS TO IT
\usepackage{caption} % DO NOT CHANGE THIS AND DO NOT ADD ANY OPTIONS TO IT
\DeclareCaptionStyle{ruled}{labelfont=normalfont,labelsep=colon,strut=off} % DO NOT CHANGE THIS
\usepackage{algorithm}
\usepackage{algorithmic}
\usepackage{xcolor}
\usepackage{multirow}
\usepackage{amsmath}

\usepackage{newfloat}
\usepackage{listings}
\floatstyle{ruled}
\newfloat{listing}{tb}{lst}{}
\floatname{listing}{Listing}

\begin{document}
% The file aaai.sty is the style file for AAAI Press 
% proceedings, working notes, and technical reports.
%
\title{Improving Non-autoregressive Generation with Mixup Training}
\author{  
    Ting Jiang\textsuperscript{\rm 1}\thanks{Work done during internship at microsoft.}, Shaohan Huang\textsuperscript{\rm 3}, Zihan Zhang\textsuperscript{\rm 4}, Deqing Wang\textsuperscript{\rm 1}\thanks{Corresponding Author},\\ Fuzhen Zhuang\textsuperscript{\rm 2}, Furu Wei\textsuperscript{\rm 3}, Haizhen Huang\textsuperscript{\rm 4}, Liangjie Zhang\textsuperscript{\rm 4}, Qi Zhang\textsuperscript{\rm 4}
}
\affiliations{
    %Afiliations

    \textsuperscript{\rm 1}SKLSDE Lab, School of Computer, Beihang University, Beijing, China \\
    \textsuperscript{\rm 2}Institute of Artificial Intelligence, Beihang University, Beijing, China\\
    \textsuperscript{\rm 3}Microsoft Research Asia \textsuperscript{\rm 4}Microsoft \\
    %If you have multiple authors and multiple affiliations
    % use superscripts in text and roman font to identify them.
    %For example,

    % Sunil Issar, \textsuperscript{\rm 2}
    % J. Scott Penberthy, \textsuperscript{\rm 3}
    % George Ferguson,\textsuperscript{\rm 4}
    % Hans Guesgen, \textsuperscript{\rm 5}.
    % Note that the comma should be placed BEFORE the superscript for optimum readability

    % 2275 East Bayshore Road, Suite 160\\
    % Palo Alto, California 94303\\
    % email address must be in roman text type, not monospace or sans serif
    % publications21@aaai.org
    \{royokong, dqwang, zhuangfuzhen\}@buaa.edu.cn\\
    \{shaohanh, zihzha, fuwei, hhuang, liazha, zhang.qi\}@microsoft.com 
    % See more examples next
    
}

\maketitle
\begin{abstract}
\begin{quote}
  While pre-trained language models have achieved great success on various natural language understanding tasks, how to effectively leverage them into non-autoregressive generation tasks remains a challenge. To solve this problem, we present a non-autoregressive generation model based on pre-trained transformer models. To bridge the gap between autoregressive and non-autoregressive models, we propose a simple and effective iterative training method called \textit{MIx Source and pseudo Target} (MIST). Unlike other iterative decoding methods, which sacrifice the inference speed to achieve better performance based on multiple decoding iterations, MIST works in the training stage and has no effect on inference time.
Our experiments on three generation benchmarks including question generation, summarization and paraphrase generation, show that the proposed framework achieves the new state-of-the-art results for fully non-autoregressive models. We also demonstrate that our method can be used to a variety of pre-trained models. For instance, MIST based on the small pre-trained model also obtains comparable performance with seq2seq models. Our code is available at \url{https://github.com/kongds/MIST}.
\end{quote}
\end{abstract}

\section{Introduction}
% Introduce pretraining models
Pre-trained models, like BERT~\cite{devlin2018bert}, UniLM~\cite{dong2019unified} and RoBERTa~\cite{liu2019roberta} have been widely applied on natural language process tasks by transferring the knowledge learned from large amount of unlabeled corpus to downstream tasks. Some pre-trained models~\cite{dong2019unified,song2019mass} also have proven effective for natural language generation tasks in autoregressive generation. However, low latency is required by an increasing number of real-world generating applications, such as online query rewriting in search engines, limiting the use of these autoregressive models~\cite{mohankumar2021diversity}.

Non-autoregressive models~\cite{gu2017non,gu2019levenshtein,ghazvininejad2019mask} are proposed to reduce latency by predicting whole sequence simultaneously. Compared with autoregressive models, non-autoregressive models can achieve more than ten times speedup at the cost of accuracy loss \cite{gu2017non}. We observe that little progress has been made to directly fine-tune an off-the-shelf pre-trained encoder model for non-autoregressive generation. How to effectively leverage them into non-autoregressive generation tasks remains a non-trivial problem.

To fully exploit the power of pre-trained models, we propose a new iterative decoding method by mixing source and pseudo target called MIST. Current iterative decoding methods such as Mask-Predict \cite{ghazvininejad2019mask} or Levenshtein transformer \cite{gu2019levenshtein} focus on improving decoding performance by multiple decoding iterations. However, these methods sacrifice the inference speed. MIST is a simple and effective iterative training strategy that works during the training stage and has no effect on inference speed. During the training stage, the model predicts the entire target sequence first, then we treat the generated target sequence as part of the source tokens and feed the source and pseudo target as source into the model.

MIST can be regarded as a dynamical data augmentation method in the training stage. Unlike the traditional data augmentation method, which needs to prepare data before training, MIST enables dynamical data augmentation in the training stage.
The term ``dynamic" in MIST refers to the fact that as model training progresses, training data will be changed per epoch. For example, MIST can ensure that the quality of the pseudo target matches the current training model and prevent model overfitting with static pseudo targets generated.
For increased data, pseudo targets can provide insight into which tokens can be successfully predicted and help models focus on what is done incorrectly. These Pseudo targets also enables the \textit{conditional dependence} to help convergence.

As our experiments show, we evaluate our method on three generation tasks including question generation, summarization and paraphrase generation. Our method achieves significant performance improvements on all tasks with the lower latency than NAT~\cite{gu2017non}.
To further evaluate the effect of our method, we compare a variety of pre-trained models including BERT~\cite{devlin2018bert}, UniLM~\cite{dong2019unified}, and MiniLM~\cite{wang2020minilm} in non-autoregressive setting.
The experiments show that our method achieves consistent gains in different generation tasks and pre-trained models.
\\ \hspace*{\fill} \\
Our contributions are listed as follows:

\begin{itemize}
\item We propose a new paradigm, adopting pre-trained encoder for non-autoregressive generation without modifying model architectures.
\item We propose a simple and novel training method to improve the performance.
\item We empirically verify the effectiveness of our method in different generation tasks and pre-trained models.
\end{itemize}
%Unlike normal iterative decoding methods, the aim of XXX is not to get better results though multiple decoding iterations, but to help our model converge. And the final model is still using one iteration to  decode.\\
%We design XXXX for both training and inference stage. For training stage, XXXX can help original model convergence and reach better results. For inference stag, model with XXXX can be a well teacher models by increasing decoding iterations.
\begin{figure*}[h]
\centering
\includegraphics[scale=0.41]{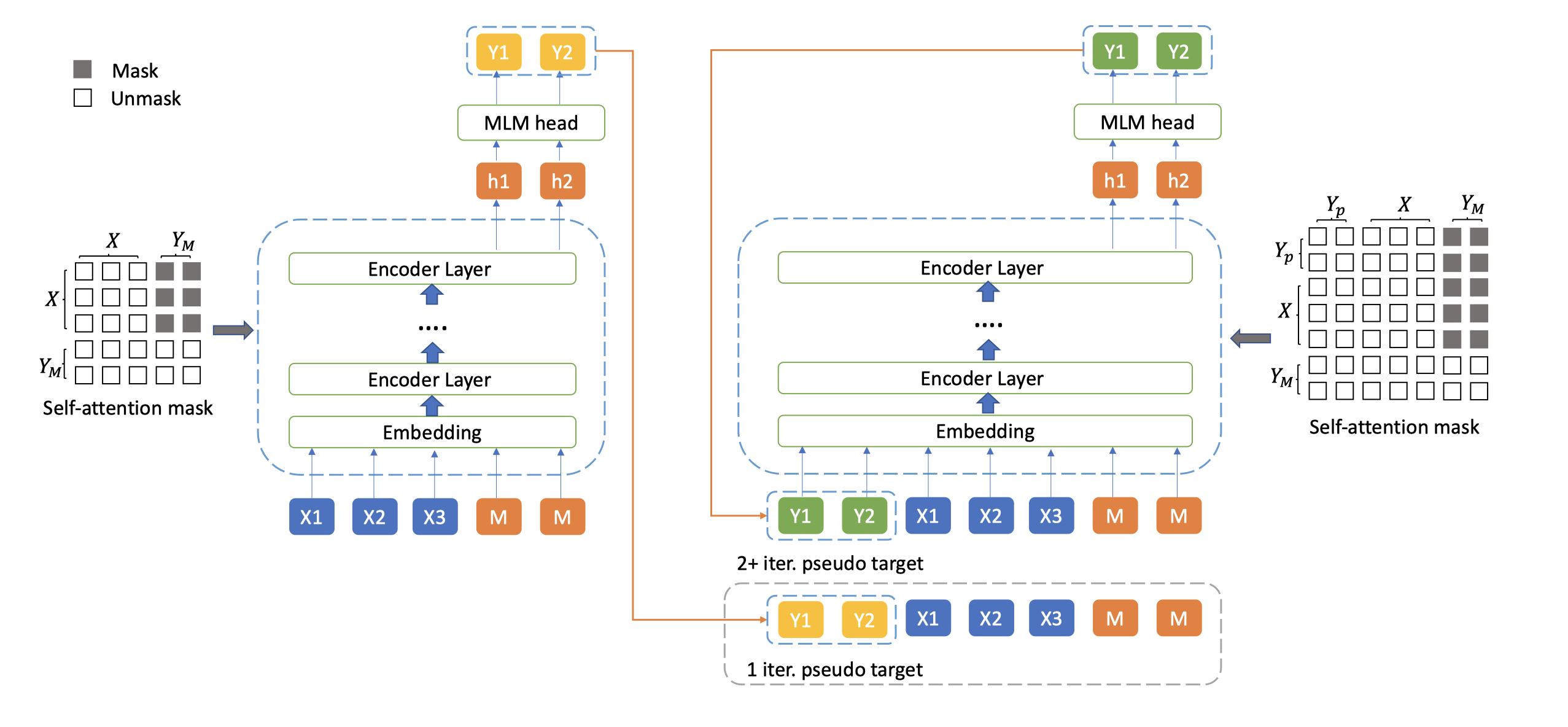}
\caption{An overview of the proposed framework}
%\caption{Overview of Mixup training. The model parameters are shared across different training iterations. We use token_type_id and attention mask to handle inputs w/o pseudo target}
\label{pic:framework}
\end{figure*}

\section{Related Work}
\paragraph{Pre-trained language models} Pre-trained language models bring significant improvement for both natural language understanding and generation tasks.
These models are trained with a large amount of unlabeled data to understand the language and improve the results on small down-stream datasets. For example,
BERT~\cite{devlin2018bert} pre-trains a encoder based model with masked language task and next sentence prediction task. It significantly improves the performance on natural language understanding tasks, but it is not suitable for generation tasks.
UniLM~\cite{dong2019unified, bao2020unilmv2} pre-trains encoder based model with three tasks: unidirectional, bidirectional, and prediction, which allow it can be fine-tuned for both natural language understanding and generation tasks. For the encoder-decoder based models~\cite{song2019mass, qi2020prophetnet, song2019mass}, these models are pre-trained with sequence-to-sequence tasks to help the down-stream generation tasks.
%The masked language modeling (MLM) as one of pre-training methods, has been widely used on many pre-trained language models

\paragraph{Non-autoregressive generation} Many works
\cite{gu2017non,kasai2020deep} have been proposed to decrease the huge latency in autoregressive generation. The most popular way is to generate tokens in parallel called non-autoregressive generation. However, these works~\cite{gu2017non,gu2020fully,qian2020glancing} mostly focus on translation, and cannot achieve reasonable results on tasks like summarization or question generation.\\ \indent BANG \cite{qi2020bang} achieves significantly improvement by bridging autoregressive and non-autoregressive generation with large scale pre-training on these tasks. For model architecture, BANG has several differences compared to previous methods: 1) BANG only uses [MASK] tokens with fixed length as the input of decoder. 2) BANG does not predict target tokens length directly, but treats the first [SEP] token as the end of sequences like autoregressive generation.

 To improve the decoding quality in non-autoregressive generation, some works~\cite{ghazvininejad2019mask,gu2019levenshtein,stern2019insertion,lee2018deterministic} use iterative decoding. The idea of iterative decoding is to refine the past decoded sequence by multiple decoding iterations. For example, Mask-Predict~\cite{ghazvininejad2019mask} first predicts target sequence non-autoregeressively, and then repeatedly mask the predicted sequence and regenerate these masked words. Levenshtein Transformer~\cite{gu2019levenshtein} reforms the generation task with the insertion and deletion operations, which refines the predicted sequence with post-editing. However, the main problem in these works is that it loses the speed advantage compared with the shallow decoder models~\cite{kasai2020deep}.

\section{Motivation}
Non-autoregressive translation has been widely studied, but the demand for efficient text generation is not only translation but also other monolingual tasks.
For example, retrieving bidding keywords with the same intent as query is a key component in search advertising system. One popular way to retrieve these keywords is rewriting the query as a generation task and then lookup the rewrite in a repository of keywords~\cite{mohankumar2021diversity}. However, the huge latency of autoregressive models makes it hard to implement online.  One solution to reduce latency is non-autoregressive generation.
%For example, non-autoregressive generation could be applied to real-time bidding keyword generation based on query in search ads system, which has strict latency requirement.

Non-autoregressive translation methods cannot be adopted into monolingual generation task directly. As the results in \cite{qi2020bang}, the non-autoregeressive translation methods show unsatisfactory results in tasks like question generation or summarization. We think two reasons cause this result.
First, the target words in translation can be aligned to the source words by using a hard alignment algorithm~\cite{brown1993mathematics}. Therefore, most non-autoregeressive translation methods~\cite{gu2017non,lee2018deterministic,guo2019non, wang2018semi} copy the source word representations to the input of the decoder. This also alleviates the multimodality problem~\cite{gu2017non} in non-autoregeressive generation.
Many non-autoregeressive translation methods are proposed for better alignment, like fertiity~\cite{gu2017non}, SoftCopy~\cite{wei2019imitation} or adding reordering module~\cite{ran2019guiding}.
%, One critical problem in non-autoregeressive generation is to construct proper decoder inputs with pre-defined length. In translation, it may initialize decoder inputs with the encoder inputs or outputs with the same methods to align the source with target like reordering the encoder output~\cite{ran2019guiding}.
However, the source and target words in monolingual generation tasks cannot be aligned directly like translation.
We cannot use copied source representations as the input of decoder to alleviate multimodality problem.
Second, the rich training data in translation allows the non-autoregressive translation methods not rely on pre-trained language models.
Simply initializing the model with BERT even harms final performance~\cite{zhu2020incorporating}. On the contrary, pre-trained language models play an important role in text generation tasks. BANG~\cite{qi2020bang} achieves the absolute improvements of 10.73, 6.39 in the overall scores of SQuAD1.1 and XSum by only using its pre-trained model.

To this end, we focus on adopting pre-trained models into non-autoregressive generation. We investigate different pre-trained models (MASS~\cite{song2019mass}, BERT~\cite{devlin2018bert}, UniLM~\cite{bao2020unilmv2}) in these tasks. We find that encoder based models are more suitable than encoder-decoder based models for these tasks. First, encoder based models achieve less latency and better performance than the encoder-decoder based models. Second, there exist many encoder based pre-trained models. We can use specific pre-trained models to satisfy our detailed requirements. For example, we can directly use a small model like MiniLM~\cite{wang2020minilm} to achieve faster decoding speed.

%To this end, we focus on the adopting the pre-trained models in these tasks. We first investigate different pre-trained models (MASS~\cite{song2019mass}, BERT~\cite{devlin2018bert}, UniLM~\cite{bao2020unilmv2}) in non-autoregeressive generation. Considering the encoder based models and encoder-decoder based models, we find the encoder based models are more suitable for this task.

\section{Proposed Framework}
In this section, we first introduce the architecture of our method and how we use it for non-autoregressive text generation.
Then we describe the implementation of the MIST. The framework is shown in Figure \ref{pic:framework}.

\subsection{Model Architecture}
 Many works \cite{he2018layer,dong2019unified} have successfully used the transformer encoder to replace the typical encoder-decoder framework in NLG tasks. 
 Inspired by these works, we directly fine-tune the pre-trained transformer encoder for non-autoregressive generation with flexible self-attention mask designs, which can efficiently use the knowledge learned from pre-training stage.
 %Inspired by these works, we directly use the original based BERT \cite{devlin2018bert} model with the MLM classification head as our architecture. The detailed information is introduced in \cite{devlin2018bert}.
 
%We use a MLM classification head to predict each tokens like MLM task in pre-training, and a simple fully connected layer is used to predict the target length. 

%%%shaohanh
Given a source sequence  $X=\left\{x_{1}, \ldots, x_{N}\right\}$, our model predicts the target $Y=\left\{y_{1}, \ldots, y_{T}\right\}$ with conditional independence, that is
\begin{equation}
    p(Y|X) = \prod_{t=1}^{T} p(y_t|X)
\end{equation}
% Formally, given a source sequence $X=\left\{x_{1}, \ldots, x_{N}\right\}$ and target sequence $Y=\left\{y_{1}, \ldots, y_{T}\right\}$.  

In the training stage, the input of our model is ``[CLS] $X$ [SEP] $Y_M$ [SEP]", where $Y_M$ considers replacing some tokens in $Y$ with [MASK]. $H_X$ and $H_Y$ are the encoded output sequences corresponding to $X$ and $Y_M$. The predicted length $\hat{T}$ is calculated by $H_X$ with a fully connected layer as multi-class classification. The tokens are predicted with MLM classification head. 

In the inference stage, we first feed our model with ``[CLS] $X$ [SEP]" to predict length $\hat{T}$, then we feed it with $\hat{T}$ length of [MASK] tokens with cached source keys and values in self-attention. By the way, we use the different token type ids to let the model distinguish the source and target tokens.

For attention mask followed \cite{dong2019unified}, we mask the source part is attention to the target part. As shown in Figure~\ref{pic:framework}, each token in the source part attends to source tokens, and each token in the target part can attends to both parts.
The source part will not be interfered by the target parts, which ensures the target length predicting is based only on source part.
In addition, we don't use positional attention, which has been widely used in other NAT models.
 %\textcolor{red}{(need one sentence to explain why not use)}.
 Although the positional attentions can improve ability to perform local reordering in \cite{gu2017non}, the time cost is still considerable compared to the final improvement.

\begin{table*}[h]
\centering
\renewcommand\arraystretch{1.1}
\begin{tabular}{llccccc}
\hline
\multicolumn{2}{c}{\multirow{2}{*}{Models}} & \multicolumn{3}{c}{SQuAD 1.1}          & \multirow{2}{*}{Iter.} & \multirow{2}{*}{Speed}\\
~                                           & ~                                      & ROUGE-L                & BLEU-4       & METEOR       & ~  & ~ \\\hline
\multirow{4}*{AR}                           & Transformer\cite{vaswani2017attention} & 29.43                  & 4.61         & 9.86         & N  & 1.0$\times$\\
~                                           &MASS\cite{song2019mass}                 & 49.48                  & 20.16        & 24.41        & N  & N/A\\
~                                           &BART\cite{lewis2020bart}                & 42.55                  & 17.08        & 23.19        & N  & N/A \\
~                                           &ProphetNet\cite{qi2020prophetnet}       & 48.00                  & 19.58        & 23.94        & N  & N/A \\\hline
\multirow{4}*{Semi-NAR}                     & InsT\cite{stern2019insertion}          & 29.98                  & 2.34         & 8.15         & 10 & 2.9$\times$\\
~                                           & iNAT\cite{lee2018deterministic}        & 32.34                  & 3.16         & 9.18         & 10 & 4.9$\times$ \\
~                                           & LevT\cite{gu2019levenshtein}           & 30.81                  & 2.68         & 9.40         & -  & 1.5$\times$ \\
~                                           & CMLM\cite{ghazvininejad2019mask}       & 29.60                  & 3.89         & 9.70         & 10 & 1.9$\times$ \\\hline
\multirow{10}*{NAR}                         & NAT\cite{gu2017non}                    & 31.51                  & 2.46         & 8.86         & 1  & 8.3$\times$\\
~                                           & BANG\cite{qi2020bang}                  & 44.07                  & 12.75        & 18.99        & 1  & *\\
~                                           & NAT MASS init                          & 46.11                  & 13.40        & 19.70        & 1  & 8.3$\times$\\
~                                           & Encoder random init\(^\dagger\)                     & 23.42                  & 1.70         & 7.37         & 1  & 8.3$\times$\\
~                                           & Encoder MinilM  init\(^\dagger\)                    & 41.70                  & 11.55        & 17.31        & 1  & 8.3$\times$\\
~                                           & Encoder BERT init\(^\dagger\)                       & 42.49                  &  11.31       &  17.53       & 1  & 8.3$\times$\\
~                                           & Encoder UnilM init\(^\dagger\)                      & 46.34                  & 16.00       & 21.10        & 1  & 8.3$\times$\\\cline{2-7}
~                                           & MIST(MiniLM)\(^\dagger\)                            & 43.87(+2.17)           & 12.98(+1.43) & 18.93(+1.62) & 1  & 8.3$\times$\\
~                                           & MIST(BERT)\(^\dagger\)                              & 43.98(+1.49)           & 12.54(+1.23) & 18.88(+1.35) & 1  & 8.3$\times$\\
~                                           & MIST(UniLM)\(^\dagger\)                             & 47.13(+0.79)           & 16.40(+0.40) & 21.68(+0.58) & 1  & 8.3$\times$\\
\hline
\end{tabular}
%\caption{Results on SQuAD 1.1. Scores of the methods marked with $\dagger$ are taken from~\cite{qi2020bang}} \label{tab:squadqg}
\caption{Results on SQuAD 1.1. \(^\dagger\): results based on our framework.} \label{tab:squadqg}
\end{table*}

\begin{table*}[h]
\centering
\renewcommand\arraystretch{1.1}
\begin{tabular}{llccccc}
\hline
\multicolumn{2}{c}{\multirow{2}{*}{Models}} & \multicolumn{3}{c}{XSum}               & \multirow{2}{*}{Iter.} & \multirow{2}{*}{Speed}\\
~                                           & ~                                      & ROUGE-1                & ROUGE-2      & ROUGE-L      & ~  & ~ \\
\hline
\multirow{4}*{AR}                           & Transformer\cite{vaswani2017attention} & 30.66                  & 10.80        & 24.48        & N  & 1.0$\times$\\
~                                           &MASS\cite{song2019mass}                 & 39.70                  & 17.24        & 31.91        & N  & N/A\\
~                                           &BART\cite{lewis2020bart}                & 38.79                  & 16.16        & 30.61        & N  & N/A \\
~                                           &ProphetNet\cite{qi2020prophetnet}       & 39.89                  & 17.12        & 32.07        & N  & N/A \\
\hline
\multirow{4}*{Semi-NAR}                     & InsT\cite{stern2019insertion}          & 17.65                  & 5.18         & 16.05        & 10 & 5.6$\times$\\
~                                           & iNAT\cite{lee2018deterministic}        & 26.95                  & 6.88         & 22.43        & 10 & 7.4$\times$ \\
~                                           & LevT\cite{gu2019levenshtein}           & 25.33                  & 7.40         & 21.48        & -  & 3.7$\times$ \\
~                                           & CMLM\cite{ghazvininejad2019mask}       & 29.12                  & 7.70         & 23.04        & 10 & 3.0$\times$ \\
\hline
\multirow{10}*{NAR}                         & NAT\cite{gu2017non}                    & 24.04                  & 3.88         & 20.32        & 1  & 12.3$\times$ \\
~                                           & BANG\cite{qi2020bang}                  & 32.59                  & 8.98         & 27.41        & 1  & * \\
~                                           & NAT MASS init                         & 27.06                  & 6.70         & 23.89        & 1  & 12.3$\times$\\
~                                           & Encoder random init\(^\dagger\)                                 & 24.36                  & 4.19         & 20.26        &1   & 12.7$\times$ \\
~                                           & Encoder MinilM  init\(^\dagger\)                                 & 21.91                  & 3.72         & 19.15        & 1  & 13.5$\times$\\
~                                           & Encoder BERT init\(^\dagger\)                                   & 28.52                  & 6.69         & 23.03        & 1  & 12.7$\times$ \\
~                                           & Encoder UnilM init\(^\dagger\)                                  & 33.66                  & 10.21        & 27.50        & 1  & 12.7$\times$ \\\cline{2-7}
~                                           & MIST(MinilM)\(^\dagger\)                            & 24.18(+2.27)           & 4.53(+0.81)  & 21.00(+1.85) & 1  & 13.5$\times$\\
~                                           & MIST(BERT)\(^\dagger\)                              & 30.50(+1.98)           & 8.00(+1.31)  & 24.83(+1.80) & 1  & 12.7$\times$\\
~                                           & MIST(UnilM)\(^\dagger\)                             & 34.63(+0.97)           & 11.29(+1.08) & 28.70(+1.20) & 1  & 12.7$\times$\\
\hline
\end{tabular}
\caption{Results on XSum. \(^\dagger\): results based on our framework.} \label{tab:xsum}
\end{table*}

\begin{table*}[h]
\renewcommand\arraystretch{1.1}
\centering
%\begin{tabular}{llp{1.67cm}<{\centering}p{1.67cm}<{\centering}p{1.67cm}<{\centering}cc}
\begin{tabular}{llccccc}
\hline
\multicolumn{2}{c}{\multirow{2}{*}{Models}} & \multicolumn{3}{c}{Quora}              & \multirow{2}{*}{Iter.} & \multirow{2}{*}{Speed}\\
~                                           & ~                                      & BLEU-1                 & BLEU-4       & METEOR       & ~  & ~ \\
\hline
\multirow{4}*{AR}                           & Transformer\cite{vaswani2017attention} & 58.57                  & 30.14        & 31.79        & N  & 1.0$\times$\\
~                                           &MASS\cite{song2019mass}                 & 60.56                  & 32.39        & 32.92        & N  & N/A\\
~                                           &BART\cite{lewis2020bart}                & 61.56                  & 31.57        & 32.42        & N  & N/A \\
~                                           &ProphetNet\cite{qi2020prophetnet}       & 62.59                  & 33.80        & 33.95        & N  & N/A \\
\hline
\multirow{4}*{Semi-NAR}                     & InsT\cite{stern2019insertion}          & 44.66                  & 21.65        & 26.82        & 10 & 2.2$\times$\\
~                                           & iNAT\cite{lee2018deterministic}        & 55.19                  & 22.72        & 27.43        & 10 & 4.3$\times$\\
~                                           & LevT\cite{gu2019levenshtein}           & 56.01                  & 27.68        & 30.56        & -  & 1.8$\times$ \\
~                                           & CMLM\cite{ghazvininejad2019mask}       & 55.63                  & 24.90        & 28.39        & 10 & 1.5$\times$ \\
\hline
\multirow{10}*{NAR}                         & NAT\cite{gu2017non}                    & 39.85                  & 9.33         & 18.90        & 1  & 5.6$\times$ \\
~                                           & NAT MASS init                           & 57.89                  & 28.16        & 30.92        & 1  & 5.6$\times$\\
~                                           & Encoder random init\(^\dagger\)                      & 48.50                  & 18.82        & 25.25        & 1  & 6.2$\times$ \\
~                                           & Encoder MinilM  init\(^\dagger\)                     & 56.87                  & 26.10        & 30.07        & 1  & 6.5$\times$ \\
~                                           & Encoder BERT init\(^\dagger\)                        & 58.31                  & 26.93        & 30.68        & 1  & 6.2$\times$ \\
~                                           & Encoder UnilM init\(^\dagger\)                       & 59.28                  & 28.51        & 31.35        & 1  & 6.2$\times$ \\\cline{2-7}
~                                           & MIST(MiniLM)\(^\dagger\)                            & 57.74(+0.87)           & 26.96(+0.86) & 30.61(+0.54) & 1  & 6.5$\times$\\
~                                           & MIST(BERT)\(^\dagger\)                              & 58.68(+0.37)           & 27.36(+0.43) & 30.90(+0.22) & 1  & 6.2$\times$\\
~                                           & MIST(UniLM)\(^\dagger\)                             & 59.65(+0.37)           & 29.00(+0.49) & 31.56(+0.21) & 1  & 6.2$\times$\\
\hline
\end{tabular}
\caption{Results on Quora. \(^\dagger\): results based on our framework.} \label{tab:quora}
\end{table*}
\subsection{MIST}
The idea of \textit{MIx Source and pseudo Target} (MIST) is simple and straightforward. The model first predicts whole target sequence with input of ``[CLS] $X$ [SEP] $Y_M$ [SEP]" and takes outputs as pseudo target $Y_p$. Then, we treat $Y_p$ as a part of source tokens and let model predict with input of ``[CLS] $Y_p$ [SEP] $X$ [SEP] $Y_M$ [SEP]", which allows model can refer to pseudo target $Y_p$.  %MIST can be regarded as an iterative decoding method in testing stage and a self sequence-level knowledge distillation method in training stage.

During the training stage, MIST maximizes two loss functions as follows \footnote{We ignore  loss function of target lenght predicting here.}:
\begin{equation}
  \mathcal{L}_{\mathrm{}}=  \mathcal{L}_{\mathrm{NAT}} +  \mathcal{L}_{\mathrm{MIST}}
%\mathcal{L}_{\mathrm{GLM}}=
\end{equation}
where $ \mathcal{L}_{\mathrm{NAT}}$ is the NAT loss function, $ \mathcal{L}_{\mathrm{MIST}}$ is the MIST loss function.
\begin{equation}
 \mathcal{L}_{\mathrm{NAT}}=-\sum_{t=1} ^{T}\log p\left(y_{t} \mid X;\theta\right)
\end{equation}

\begin{equation}
  \begin{aligned}
    \mathcal{L}_{\mathrm{MIST}}&=-\sum_{t=1} ^{T}\log p\left(y_{t} \mid Y_p, X;\theta\right)\\
    &=-\sum_{t=1} ^{T}\log p\left(y_{t} \mid G(X;\theta), X;\theta\right)
  \end{aligned}
\end{equation}
where $\theta$ denotes the parameter set of encoder model, $Y_p=G(X;\theta)$ denotes the pseduo target $Y_{p}$ of $X$ generated by $\theta$. 
% $ \mathcal{L}_{\mathrm{NAT}}$ is the NAT loss function. $\mathcal{L}_{\mathrm{MIST}}$ is the loss function of MIST. 

The $\mathcal{L}_{\mathrm{MIST}}$  can be regarded as the data augmentation method, which regenerates the  target sequence by referencing the previous pseudo target $Y_{p}$ as the part of source sequence.
Due to the fact that we can't copy source representation as the decoder input in monolingual text generation task, the multimodality problem in NAT can't be solved by using the latent variable to model the non-determinism~\cite{gu2017non}.
MIST is a solution to tackle the multimodality problem by converting the $p(y_{t}|X)$ to $p(y_{t}|Y_{p},X)$, which enables the \textit{conditional dependence} during training. Each predicted token's distribution depends on the source sequence $X$ and pseduo target targets $Y_{p}$ to help the model approximate true target distribution.

During the inference stage, we mainly focus on single-pass parallel generation.
With the help of MIST training, the model can achieve better performance on single-pass parallel generation.
Moreover, MIST training also provides a novel iterative decoding method  by feeding the model with mixing pseudo target and source sequence to achieve better decoding quality. In ablation study, we find the MIST iterative decoding achieves better performances than Mask-Predit~\cite{ghazvininejad2019mask}.

\section{Experiment}
In this section, we first introduce the datasets and baselines. Then we report the results of MIST on three different pre-trained models.

\subsection{Dataset}
We conduct our non-autoregressive generation experiments on three popular generation datasets:

\paragraph{SQuAD 1.1} SQuAD 1.1~\cite{rajpurkar2016squad} is a reading comprehension dataset, and we follow the setup~\cite{zhou2017neural, zhao2018paragraph} as an answer-aware question generation task.

\paragraph{XSum} XSum~\cite{narayan2018don} is a dataset for evaluation of abstractive single-document summarization systems, which has 227K online articles from the British Broadcasting Corporation (BBC).
% And the source input is the article, the target output is the summary.

% After the pre-processing, the dataset has 98K <answer, passage, question> data triples. And we feed model with <answer [SEP] passage>.

\paragraph{Quora} Quora\footnote{\url{https://quoradata.quora.com/First-Quora-Dataset-Release-Question-Pairs}} has more than 400K potential question duplicate pairs.  We follow the setup in \cite{gupta2018deep,huang2019dictionary} and use the duplicate question pairs as a paraphrase generation task.
% Each line contains IDs for each question in the pair, the full text for each question, and a binary value that indicates whether the line truly contains a duplicate pair.
% The average number of words in source and target part are 11.19 and 11.17 respectively.

\subsection{Baselines}
%In the experiments, we follow the baselines in~\cite{qi2020bang}. Moreover, we also report the NAR encoder decoder model inited with pre-trained model like MASS
%\textcolor{red}{maybe we can call ours as mist (including dat + kd), and show -dat; -kd; - dat -kd}
We follow the baselines in~\cite{qi2020bang} and use autoregressive (AR), non-autoregeressive (NAR) and semi-NAR baselines.

For AR baselines, we use four models including three pre-trained generation models. (1) Vanilla Transformer~\cite{vaswani2017attention}. (2) MASS~\cite{song2019mass}, a sequence-to-sequence pre-trained model where the encoder takes the masked sequence as input and the decoder predicts masked tokens auto-regressively. (3) BART~\cite{lewis2020bart}, similar frameworks with MASS, which trains the model as a denoising autoencoder. (4) ProphetNet~\cite{qi2020prophetnet}, a new sequence-to-sequence pre-trained model, which is pre-trained by predicting future n-gram based on n-stream self-attention mechanism.

For NAR baselines, we compare to NAT~\cite{gu2017non}, BANG~\cite{qi2020bang}, NAT initialized from MASS and encoder based methods. For BANG, it is a new NAR pre-trained model, which achieves significant improvements compared to previous NAR models. For NAT initialized from MASS,
we initialize the parameters of NAT transformer with the pre-trained MASS model and change the undirectional self attention mask in decoder to bidirectional. We also add the length predicting head in encoder.
For encoder based methods, we use three pre-trained models \textit{unilm1.2-base-uncased}~\cite{bao2020unilmv2}, \textit{bert-base-uncased}~\cite{devlin2018bert} and \textit{MiniLMv1-L12-H384-uncased}~\cite{wang2020minilm} to initialize the encoder. Moreover, we also add results of random initialized encoder.

For semi-NAR baselines, we compare with several typical semi-NAT models: (1) iNAT~\cite{lee2018deterministic}, a conditional NAR model based on iterative refinement mechanism. (2) CMLM~\cite{ghazvininejad2019mask}, a conditional masked NAR model based on mask-predict. (3) InsT~\cite{stern2019insertion}, which iteratively inserts tokens in the sequence during decoding. (4) LevT~\cite{gu2019levenshtein}, which models the generation as multi-step insertion and deletion operations based on imitation learning.

\subsection{Experimental Setting}
We use the BERT-base model ($n_{layers}=12, n_{heads}=12, d_{hidden}=768$) and BERT-small model ($n_{layers}=12, n_{heads}=12, d_{hidden}=384$) as our backbone based on huggingface transformers \cite{wolf2020transformers}.
The weight of model is respectively initialized  by \textit{unilm1.2-base-uncased}~\cite{bao2020unilmv2}, \textit{bert-base-uncased}~\cite{devlin2018bert} and \textit{MiniLMv1-L12-H384-uncased}~\cite{wang2020minilm}.
We use GLAT \cite{qian2020glancing} in training stage and set mask raito $f_{ratio}=0.5$ in encoder based baselines and MIST.
The training hyper-parameters are: batch size 64, learning rate 3e-5, warm up steps 500.
The maximum input length is set to 464, and the maximum output length is set to 48. We measure the validation BLEU-4 scores for every 5,000 updates, and choose the best checkpoints.  We don't use sequence-level knowledge distillation followed by BANG. The latency is evaluated by using a single Nvidia V100 GPU and setting batch size as 1.

%For knowledge distillation, we use the same training hyper-parameters and model architecture as teacher model. And the sequence-level knowledge distillation is generated by 6 decoding iterations based on MIST in semi-NAR way. By the way, the teacher model is equivalent to our model with DAT. The purpose of DAT is to let the teacher model learn how to iterate decoding in MIST. Furthermore, we will investigate the influence of teacher models in ablation study.

For SQuAD1.1 and XSum, the AR baseline results are collected from GLGE \cite{liu2020glge}. The NAR and semi-NAR baseline results are directly referred from \cite{qi2020bang}. For Quora, we use sentencepiece~\cite{kudo2018sentencepiece} to generate sub-words as tokens for no pre-trained models. For inference, we set the beam size as 4, length penalty as 1.0 for AR baselines and set the maximum iteration steps as 10 for all semi-NAR baselines.
We also remeasure the latency for three datasets in our hardware environment based on the Fairseq \cite{ott2019fairseq} v0.10.2 codebase. Furthermore, we find the speedup between AR and NAR baselines is lower than \cite{qi2020bang}. We guess that it is caused by different hardware environments.

\subsection{Main Results}
The results of question generation task, summarization task and paraphrase generation task are shown in Table \ref{tab:squadqg}, Table \ref{tab:xsum} and Table \ref{tab:quora}, respectively. Moreover, we report the NAR models initialized from BERT, UniLM, MiniLM and MASS as our baselines.

Our encoder based models achieve significant performance improvements on all tasks with the lower latency than NAT \cite{gu2017non}.
Despite the vast differences in these datasets, MIST can successfully work on all of them. Furthermore, the improvement of MIST on various pre-trained models is constant and significant.
Compared to BANG \cite{qi2020bang}, which uses powerful large-scale pre-training for NAR generation, our method still achieves better results by adopting off-the-shelf pre-trained model.
Due to the code of BANG not being released yet, we do not measure the speedup in three datasets and the result in Quora dataset.

We observe that in one batch size, the speedup in MiniLM is not obvious compared to the base size models. When use the 16 batch size, MiniLM can be twice as fast as the base size models.

From Table \ref{tab:squadqg}, our method achieves the best result among semi-NAR and NAR models. Meanwhile, our method even outperforms the AR transformer. On the other hand, the baselines based on pre-trained models significantly outperform the random initialized baselines, which achieve less than 5\% BLEU-4. Among AR pre-trained models, MASS achieves the best results on SQuAD1.1. In NAR pre-trained models, MIST(UniLM) achieves absolute improvements of 3.65 in BLEU-4 compared to the powerful per-trained NAR model BANG.

For the summarization task in Table \ref{tab:xsum}, our method outperforms the other semi-NAR and NAR models. Due to the longer source and target sequences, the summarization task is more challenging for non-autoregressive setting. MIST is more effective and  brings a significant boost to this task.
%The results of our model are also better than AR transformer.% in ROUGE-1, ROUGE-2 and ROUGE-L.

For the paraphrase generation task in Table \ref{tab:quora}, pre-training has a smaller impact than other two datasets. Meanwhile, we find that the semi-NAR baselines can achieve better result than NAT. For example, the semi-NAR models like LevT can achieve improvements of 18.35 in BLEU-4 compared with NAT. However, the speedup of semi-NAR models is limited compared to AR transformer. Furthermore, we observe that our method achieves more speedup than NAT, and we can achieve 6.2 speedup times compared to 5.6 speedup times in NAT. The main reason may be that our method uses self-attention layer to replace both self-attention layer and cross-attention layer in the NAT decoder part.
\\ \hspace*{\fill} \\

\section{Ablation Study}

\subsection{Influence of Mixup Source and Target Methods}
To study the influence of MIST training, we propose a static pseudo targets mix training method (static mixing), which generates pseudo targets from the trained models, and then training with both mixing data and original data.
The results of three methods in SQuAD1.1 and XSum are shown in Table \ref{tab:da-compare-squad} and Table \ref{tab:da-compare-xsum}. 
For static mixing, it improves the performance in SQuAD1.1, but harms the performance in XSum. However, MIST improve performance in both SQuAD1.1 and XSum compared to no mixing. We also observe that the gains of MIST exceed static mixing.

\begin{figure}[h]
\centering
\includegraphics[scale=0.33]{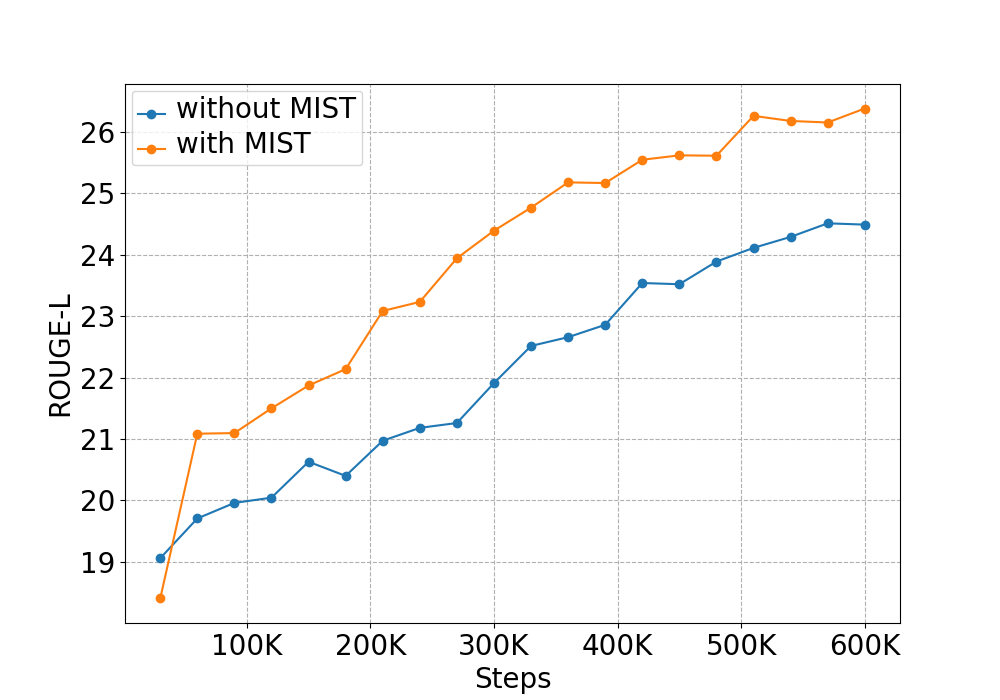}
\caption{Influence of the MIST on model convergence in XSum}
\label{pic:xsum_kd_da_compare}
\end{figure}

We also analyze the influence of MIST on model convergence and check the ROUGE-L score on validation set at every 30K steps during training in XSum. As the Figure \ref{pic:xsum_kd_da_compare} shown, we observe that MIST can lead to faster convergence compared to the original model. Moreover, MIST outperforms the original model with only half training steps.
\begin{table}[h]
\small
\centering
\begin{tabular}{lccc}
\hline 
 & ROUGE-L & BLEU-4 & METEOR \\
\hline
No Mixing& 45.67 & 15.33 & 20.89 \\
Static Mixing & 45.74 & 15.89 & 21.29 \\
MIST & 46.49 & 16.10 & 21.59 \\
\hline 

\end{tabular}
\caption{Influence of different mixup methods in SQuAD1.1} \label{tab:da-compare-squad}
\end{table}

\begin{table}[h]
\small
\centering
\begin{tabular}{lccc}
\hline 
     & ROUGE-1 & ROUGE-2 & ROUGE-L \\
\hline
 No Mixing & 33.66 & 10.21 & 27.50 \\
Static Mixing & 32.46 & 9.41 & 26.42 \\
MIST & 34.63 & 11.29 & 28.70 \\
\hline 

\end{tabular}
\caption{Influence of different mixup methods in XSum} \label{tab:da-compare-xsum}
\end{table}

%\subsection{Influence of the MIST on token repetitions}
%To analyze the influence of MIST training, we compute the token repetition rate in each dataset in Table \ref{tab:kd-da-repeat}. We find the MKD and DAT can effectively reduce the repetition rate, especially in XSum, which can be reduced by more than 17\% by simply using MKD and DAT. For SQuAD1.1 and Quora, the MKD and DAT can also help model generate sequence with low repetition rate, even though the original repetition rate is already very low.
%\begin{table}[h]
%\small
%\centering
%\begin{tabular}{lccc}
%\hline
%& XSum & SQuAD1.1 & Quora \\
%\hline
%our model & 34.86 & 10.74 & 6.93 \\
%\hspace{0.5em} +DAT  & 30.26 & 9.86 & 6.81\\
%\hspace{0.5em} +MKD  & 17.81 & 8.74 & 6.16\\
%\hspace{0.5em} +DAT +MKD & 17.76 & 8.24 & 6.06\\
%\hline
%\end{tabular}
%\caption{Influence of the MKD and DAT in token repetitions} \label{tab:kd-da-repeat}
%\end{table}
%
%We also analyze the influence of MKD and DAT on model convergence, and we check the ROUGE-L of validation set at every 30K steps during training in XSum. As the Figure \ref{pic:xsum_kd_da_compare} shown, we observe both MKD and DAT can help model convergence. We also find that MKD can shorten the overall training steps. By combining MKD and DAT, the results on 30K steps can exceed the 600k steps, and it only needs 180K steps to reach the best result.

%\begin{figure}[h]
%\centering
%\includegraphics[scale=0.33]{xsum_kd_da_compare_2.png}
%\caption{Influence of the MKD and DAT in XSum}
%\label{pic:xsum_kd_da_compare}
%\end{figure}

\subsection{MIST vs Mask-Predict}
We also evaluate the effectiveness of MIST in iterative decoding compared to Mask-Predict~\cite{ghazvininejad2019mask}. For fair comparison, Mask-Predict is implemented based on our method, and we use same linear decay in \cite{ghazvininejad2019mask} to decide the number of masked tokens in every iteration. We have shown the result of 1, 3, 6 decoding iterations respectively on SQuAD1.1, XSum and Quora in Table \ref{tab:mask_predict_MIST}. We choose BLEU-4 as the evaluation metrics in SQuAD1.1 and Quora, and ROUGE-2 in XSum.

\begin{table}[h]
\centering
\begin{tabular}{lcccc}
\hline 
Method       & Iter. & SQuAD1.1 & XSum  & Quora \\
\hline
MIST         & 1     & 16.09    & 11.17 & 28.95 \\
Mask-Predict & 1     & 16.09    & 11.17 & 28.95 \\
\hline 
MIST         & 3     & 18.12    & 12.46 & 29.77 \\
Mask-Predict & 3     & 16.73    & 11.27 & 27.70 \\
\hline 
MIST         & 6     & 18.27    & 12.79 & 29.77 \\
Mask-Predict & 6     & 16.57    & 11.70 & 27.69 \\
\hline 
\end{tabular}
\caption{Iterative decoding results of MIST and Mask-Predict} \label{tab:mask_predict_MIST}
\end{table}

As Table \ref{tab:mask_predict_MIST} shows, MIST outperforms  Mask-Predict as iterative decoding methods in the encoder based model.
Meanwhile, we find that Mask-Predict cannot improve the result in Quora. The improvement obtained by more iterations are not consistent in Mask-Predict. For example, the result of 6 iterations is worse than 3 iterations in SQuAD1.1. As the number of iterations increases, the gain of MIST is more stable than Mask-Predict.

%\subsection{Influence of two loss functions in MIST  }
%We also check the influence of two loss functions $ \mathcal{L}_{\mathrm{NAT}}$, $ \mathcal{L}_{\mathrm{MIST}}$  in MIST. We have

\section{Discussion}
\subsection{Comparing BANG with MIST}
In this section, we compare the state-of-the-art method BANG with MIST.
BANG is a pre-trained model, which bridges the gap between AR and NAR.
MIST is a training method to adopt encoder based models in NAR.
One advantage of MIST is that it can directly adopt pre-trained models without modification in model architectures, which saves the large computational resources in pre-training and allows it to use existing pre-trained models more flexibly.

Moreover, we also compare the gains of two methods. For BANG, we calculated the gains from NAR pre-trained model, which is reported in \cite{qi2020bang}.
We found our method is comparable to the BANG with only MIST training in fine-tuning stage. Our method can successfully adopt the NLU pre-trained models like BERT to the powerful NAT models.
%Especially in BERT and MiniLM, MIST significantly shorten the gap between the
%By the way, MIST improves the results consistently in all three different pre-trained models.

\begin{table}[h]
\small
\centering
\renewcommand\arraystretch{1.2}
\begin{tabular}{lcccc}

\hline
              & ROUGE-L      & BLEU-4       & METEOR       \\
\hline
BANG          & 44.07(+2.92) & 12.75(+1.37) & 18.99(+1.36)\\
MIST(MiniLM)  & 43.87(+2.17) & 12.98(+1.43) & 18.93(+1.62)\\
MIST(BERT)    & 43.98(+1.49) & 12.54(+1.23) & 18.88(+1.35)\\
MIST(UniLM)   & 47.13(+0.79) & 16.40(+0.40) & 21.68(+0.58)\\
\hline
\hline
              & ROUGE-1      & ROUGE-2      & ROUGE-L       \\
\hline
BANG          &  32.50(+2.07)& 8.92(+1.26)  & 27.35(+1.85)\\
 MIST(MinilM) & 24.18(+2.27) & 4.53(+0.81)  & 21.00(+1.85) \\
 MIST(BERT)   & 30.50(+1.98) & 8.00(+1.31)  & 24.83(+1.80) \\
 MIST(UnilM)  & 34.63(+0.97) & 11.29(+1.08) & 28.70(+1.20) \\
\hline
\end{tabular}
\caption{Compring between BANG and MIST on SQuAD1.1 and XSum} \label{tab:bang}
\end{table}

\section{Conclusion}

In this paper, we proposed a new paradigm to adopt pre-trained encoders to NAR tasks. Our method uses a transformer encoder as model architecture with a new iterative decoding method called MIST to improve the generation quality without additional latency in decoding.
Our method achieves state of the art results among the NAR and semi-NAR models on three different NLG tasks.
We also demonstrated that our method can successfully utilize the pre-trained models to achieve better results than large-scale NAR pre-trained models like BANG.
%Compared to BANG, our framework can achieve better results without large-scale NAR pre-training, which can prove our framework can efficiently adopt the pre-trained model to NAR tasks.

% Entries for the entire Anthology, followed by custom entries
\bibliography{custom}

%\appendix

%\section{Example Appendix}
%\label{sec:appendix}

%This is an appendix.

\end{document}